\documentclass[11pt]{article}

\usepackage[utf8]{inputenc} 
\usepackage[T1]{fontenc}    
\usepackage{lmodern}
\usepackage[hidelinks]{hyperref}       
\usepackage{booktabs}       
\usepackage[numbers]{natbib}
\usepackage{nicefrac}       
\usepackage{microtype}      

\usepackage[linesnumbered, ruled, vlined, noend]{algorithm2e}
\usepackage{algorithmic}
\usepackage{amsfonts,amsmath,amsthm,amssymb,bm}
\usepackage[title]{appendix}
\usepackage{courier}
\usepackage[usenames,dvipsnames]{color}
\usepackage{enumitem}
\usepackage[margin=1in]{geometry}
\usepackage{url}

\input{./Definitions}

\allowdisplaybreaks

\title{Structural Conditions for Projection-Cost Preservation via Randomized Matrix Multiplication}

\author{
Agniva Chowdhury%
\thanks{Department of Statistics, Purdue University,
West Lafayette, IN. Emails:\,\url{{chowdhu5, jiaseny}@purdue.edu}.
Both authors contributed equally to this work.}
\and
Jiasen Yang%
\footnotemark[1]
\and
Petros Drineas%
\thanks{Department of Computer Science, Purdue University,
West Lafayette, IN. Email:~\url{pdrineas@purdue.edu}.}
}

\date{}

\begin{document}

\maketitle


\begin{abstract}
\noindent \emph{Projection-cost preservation} is a low-rank approximation guarantee which ensures that the cost of any rank-$k$ projection can be preserved using a smaller sketch of the original data matrix. We present a general structural result outlining four sufficient conditions to achieve projection-cost preservation. These conditions can be satisfied using tools from the Randomized Linear Algebra literature.

\end{abstract}

{\small
\emph{Keywords:}~%
Projection-cost preservation, Randomized linear algebra, Matrix sketching, Leverage scores

\emph{Mathematics Subject Classification:}~%
15A23,
15A45,
65F30,
68W20,  
68W25  
}


\section{Introduction}

Projection-cost preservation is a low-rank approximation guarantee which ensures that the cost of any $k$-rank projection can be preserved using a smaller sketch of the original data matrix. It has recently emerged as a fundamental principle in the design and analysis of sketching-based algorithms for common matrix operations that are critical in data mining and machine learning.

Prior to introducing the formal definition for cost-preserving projections, we state the general \emph{constrained low-rank approximation} problem. For any matrix $\Ab\in\RR{n}{d}$ and a set $\Omegab$ of orthogonal projection matrices $\Pb\in\RR{d}{d}$ with $\rank(\Pb)=k$, we seek the minimizer of the following optimization problem:
\begin{flalign}
\Pb^{*}=\argmin_{\Pb\in\Omegab}\|\Ab-\Ab\Pb\|_F^2\label{eq:cons}\,.
\end{flalign}
Here, the term $\|\Ab-\Ab\Pb\|_F^2$ is called the \textit{cost} of projection $\Pb$; recall that $\|\Xb\|_F^2 = \sum_{i,j} \Xb_{ij}^2$. Two simple examples will help clarify the importance of the above formulation. First, let $\Omegab$ be the set of all rank-$k$ orthogonal projection matrices: in this case, the problem of eqn.~\eqref{eq:cons} is equivalent to finding the best rank-$k$ approximation to $\Ab$, which can be computed via the Singular Value Decomposition (SVD) in polynomial time. Second, let $\Ab$ be a data matrix whose columns represent $d$ points in $\mathbb{R}^n$ and let $\Omegab$ be the set of all orthogonal rank-$k$ projection matrices of the form $\Pb=\Xb\Xb^\ts$. Here, $\Xb\in\RR{d}{k}$ is the \textit{rescaled} cluster membership matrix, \ie, $\Xb_{ij}=\nicefrac{1}{\sqrt{s_j}}$ if the $i$-th column of $\Ab$ belongs to the $j$-th cluster and zero otherwise; $s_j$ is the size of $j$-th cluster.%
\footnote{Note that every row of $\Xb$ has exactly one non-zero element and its columns are pairwise orthogonal and normal.}
In this case, the constrained low-rank approximation problem of eqn.~(\ref{eq:cons}) is equivalent to the well-known $k$-means clustering problem~\cite{DFKVV04}.

The above discussion shows that, depending on the set $\Omegab$, the optimization problem of eqn.~(\ref{eq:cons}) can be easy or very hard to solve exactly. Indeed, the low-rank approximation problem can be solved in polynomial time via the SVD, whereas the $k$-means clustering problem is NP-hard~\cite{Aloise2009} and a polynomial time algorithm is unlikely. The projection cost preservation formulation of the optimization problem of eqn.~(\ref{eq:cons}) replaces the full matrix $\Ab$ by a smaller sketch $\widetilde{\Ab}\in\RR{s}{d}$, with $s\ll n$, in order to reduce the solution cost. The following definition first appeared in~\cite{CohEldMusMusetal15,Musco15}; see Section~\ref{sec:prior} for a discussion of prior work.%
\footnote{The original definitions of~\cite{CohEldMusMusetal15,Musco15} included a non-negative, fixed constant $c$ as an additive term, which does not add any generality in our setting.}

\begin{definition}\label{def:pcps} Let $\Ab \in \mathbb{R}^{n \times d}$ be the input matrix and let $\Wb\in\RR{s}{n}$ with $s\ll n$ be a sketching matrix. The matrix $\Wb\Ab$ is a rank-$k$ \emph{projection-cost preserving sketch} of $\Ab$ with error $\ve \in [0,1]$ if it satisfies
\begin{flalign}
	(1-\ve)\|\Ab-\Ab\Pb\|_F^2\le\|\Wb\Ab-\Wb\Ab\Pb\|_F^2\le(1+\ve)\|\Ab-\Ab\Pb\|_F^2\label{eq:pcps}\,,
\end{flalign}
for all rank-$k$ projection matrices $\Pb\in\RR{d}{d}$ $(1 \leq k < d)$.
\end{definition}
\noindent In words, the so-called \emph{projection-cost preserving sketch} $\Wb\Ab$ can replace the original matrix $\Ab$ in the optimization problem of eqn.~(\ref{eq:cons}) with a small loss in accuracy (see Lemma~3 of \cite{CohEldMusMusetal15}) and thus one can solve the \textit{sketched} problem instead of the original problem. We will now slightly manipulate Definition~\ref{def:pcps} by rewriting the rank-$k$ projection matrix $\Pb\in\RR{d}{d}$ as follows: let $\Xb\in\RR{d}{(d-k)}$ be a matrix whose columns form a basis for the subspace that is \textit{orthogonal} to the subspace spanned by  $\Pb$. Thus, $\Pb=\Ib_d-\Xb\Xb^\ts$ and $\Xb^\ts\Xb=\Ib_{d-k}$ ($\Ib$ is a square identity matrix of appropriate dimensions). We can now rewrite the exact and approximate cost of the projection $\Pb$ as follows:
\begin{flalign*}
  \|\Ab-\Ab\Pb\|_F^2 &=\|\Ab\Xb\Xb^\ts\|_F^2=\|\Ab\Xb\|_F^2,\ \mbox{and} \\
  \|\Wb\Ab-\Wb\Ab\Pb\|_F^2 &=\|\Wb\Ab\Xb\Xb^\ts\|_F^2=\|\Wb\Ab\Xb\|_F^2.
\end{flalign*}
The final equalities in both derivations follow from the unitary invariance of the Frobenius norm. We can now state our (equivalent) definition of cost-preserving projections.
\begin{definition}[Cost-preserving projections]\label{def:pcps2}
Let $\Ab \in \mathbb{R}^{n \times d}$ be the input matrix and let $\Wb\in\RR{s}{n}$ with $s\ll n$ be a sketching matrix. $\Wb\Ab$ is a rank-$k$ projection-cost preserving sketch of $\Ab\in\RR{n}{d}$ with error $\ve \in [0,1]$, if it satisfies
\begin{flalign}
	\abs{\|\Wb\Ab\Xb\|_F^2-\|\Ab\Xb\|_F^2}\le\ve\,\|\Ab\Xb\|_F^2\label{eq:pcps2}\,,
\end{flalign}
for all matrices $\Xb\in\RR{d}{(d-k)}$ such that $\Xb^\ts\Xb=\Ib_{d-k}$ $(1 \leq k < d)$.
\end{definition}
\noindent Building upon the above definition, our main contribution is a general, structural result (Theorem~\ref{thm:structural}) presenting four sufficient conditions that a sketching matrix $\Wb$ should satisfy in order to guarantee that the sketched matrix $\Wb\Ab$ is a cost-preserving projection. The proposed sufficient conditions all boil down to sketching-based matrix multiplication (see Section~\ref{sec:mm} for a review), a fundamental and well-studied primitive of Randomized Linear Algebra (RLA). Such structural results have been of paramount importance in the RLA community, as they typically decouple the linear-algebraic component of a problem from the randomized algorithms that are employed to satisfy the structural conditions. See~\cite{Halko2011,Drineas11,DM2016chapter} for similar structural results for a variety of linear algebraic problems for which randomized algorithms have been designed, including $\ell_2$ regression, SVD approximation, the Column Subset Selection Problem, etc. In Section~\ref{sxn:examples}, we instantiate Theorem~\ref{thm:structural} to show how different constructions of the sketching matrix $\Wb$ satisfy the structural conditions of our theorem. We hope that this linear-algebraic exposition of the cost-preserving projection problem will help bring it to the forefront of the linear algebra community and stimulate further research on this fundamental
problem.

\subsection{Related Work}\label{sec:prior}
The importance of cost-preserving projections was first recognized by~\cite{CohEldMusMusetal15,Musco15}, who coined the aforementioned term for the problem of Definition~\ref{def:pcps2}. Their work provided several ways to construct provably accurate projection-cost preserving sketches and also demonstrated their applicability to constrained low-rank approximation problems, such as $k$-means clustering. Their work recognized the importance of structural results for cost-preserving projections and actually presented a related structural theorem, which is considerably more involved and complicated than our Theorem~\ref{thm:structural}. In a more recent paper~\cite{CohMusMus17}, the authors showed that ridge-leverage score sampling also satisfies the cost-preserving projection guarantee. Interestingly, the results of~\cite{CohEldMusMusetal15,Musco15} and~\cite{CohMusMus17} are quite independent and different from each other in terms of proof strategies, at least to the best of our understanding. A major motivation of our work was the unification of these two seemingly different approaches for cost-preserving projections using the same structural result.

We do note that prior to~\cite{CohEldMusMusetal15,Musco15}, the idea of projection-cost preservation was also discussed in~\cite{FelSchSoh13} (see their Definition~2) and it was also implicit in~\cite{DFKVV04,BZMD2015}. Cost-preserving projections have also been connected with the construction of \emph{coresets}%
\footnote{In words, coresets are small sets of points that approximate the shape and properties of a larger point-set.}
in machine learning, computational geometry, and theoretical computer science. We refer the interested reader to~\cite{bachem17,FelSchSoh13,FT15,FelVolRus16} for detailed discussions. In particular,~\cite{FelVolRus16} addressed the existence and efficient construction of coresets using a definition that is almost identical to Definition~\ref{def:pcps2}.

\section{Notation}\label{sec:prelim}

Given a matrix $\Ab\in\RR{n}{d}$, let $\Ab_{i*}$ denote the $i$-th row of $\Ab$ as a row vector and let $\Ab_{*i}$ denote the $i$-th column of $\Ab$ as a column vector. Let the (thin) SVD of $\Ab$ be $\Ab = \Ub_{n\times r} \Sigmab_{r\times r} \Vb^\ts_{d\times r}$; subscripts denote matrix dimensions and $r$ denotes the rank of the matrix $\Ab$. It is well-known that $\Ub^\ts\Ub = \Vb^\ts\Vb = \Ib_r$ and $\Sigmab = \diag\{\sigma_1,\dots, \sigma_r\}$ consists of the non-zero singular values of $\Ab$ sorted in non-increasing order, $\sigma_1\ge \dots\ge \sigma_r > 0$. Let $\Ab_m$ be the best rank-$m$ approximation to $\Ab$ and let $\Ab_{m, \perp}=\Ab-\Ab_m$. We will use the slightly non-standard notation $\Sigmab_m \defeq \Diag{\sigma_1, \dots, \sigma_m, 0, \dots, 0}$ to denote the $r \times r$ diagonal matrix whose top $m$ entries are equal to the top $m$ singular values of $\Ab$ and the bottom $r-m$ entries are set to zero. Similarly, $\Sigmab_{m,\perp} \defeq \Diag{0, \dots, 0, \sigma_{m+1}, \dots, \sigma_r}$ is the $r \times r$ diagonal matrix whose top $m$ entries are set to zero and the bottom $r-m$ entries are set to the bottom $r-m$ singular values of $\Ab$. Clearly, $\Sigmab = \Sigmab_m + \Sigmab_{m,\perp}$, $\Ab_{m}=\Ub\Sigmab_m\Vb^\ts$, and $\Ab_{m,\perp}=\Ub\Sigmab_{m,\perp}\Vb^\ts$.

We will make frequent use of matrix norms, norm inequalities, and matrix trace properties; we refer the reader to Chapter~2 of~\cite{DM2018} for a quick introduction. We do note the strong submultiplicativity property of the Frobenius norm, namely that for any two matrices $\Ab$ and $\Bb$ of suitable dimensions,
\[
\nbr{\Ab\Bb}_F \leq \min\{ \nbr{\Ab}_2 \cdot \nbr{\Bb}_F,\, \nbr{\Ab}_F \cdot \nbr{\Bb}_2\}.
\]
An important tool in our analysis will be \emph{von Neumann's trace inequality}; recall that the trace of a square matrix $\Ab$, denoted $\tr(\Ab)$, is the sum of its diagonal entries.
\begin{proposition}[Von Neumann's trace inequality~\cite{Mirsky1975}]
\label{lem:von-neumann}
For any matrices $\Ab,\Bb\in\RR{n}{n}$ with singular values
$\sigma_1(\Ab) \ge \sigma_2(\Ab) \ge \dots \ge \sigma_n(\Ab)$
and
$\sigma_1(\Bb) \ge \sigma_2(\Bb) \ge \dots \ge \sigma_n(\Bb)$
respectively,
\begin{equation*}
  |\tr(\Ab\Bb)| \le \sum_{i=1}^n \sigma_i(\Ab) \, \sigma_i(\Bb) \,.
\end{equation*}
\end{proposition}
In Section~\ref{sxn:examples}, we will make frequent use of a fundamental result from probability theory, known as \emph{Markov's inequality}. Let $X$ be a random variable assuming non-negative values with expectation $\Ex{X}$. Then, for any $t > 0$,
\[
\pr{X \ge t \cdot \Ex{X}} \le \frac{1}{t} \,.
\]
%
We will also need the so-called \emph{union bound}: given a set of random events ${\cal E}_1,{\cal E}_2,\ldots,{\cal E}_{t}$ holding with respective probabilities $p_1,p_2,\ldots,p_t$, the probability that at least one of these events hold (\ie, the probability of the union of these events) is upper-bounded by $\sum_{i=1}^t p_i$.


\section{Our Structural Result}\label{sxn:structural}
We now describe our main structural result for cost-preserving projections (see Definition~\ref{def:pcps2}). Our structural result connects cost-preserving projections with sketching-based matrix multiplication, a well-studied primitive in the Randomized Linear Algebra community.

Prior to presenting our result we define the diagonal matrix $\Sreg\in\RR{r}{r}$ as
\begin{equation}\label{eqn:pd1}
 \Sreg=\diag\{d_1,d_2,\ldots,d_m,\ldots,d_q,0,\ldots,0\},
\end{equation}
with $d_1\geq d_2 \geq \ldots \geq d_m \geq \ldots \geq d_q >0$. As we will discuss in more detail below, $q$ and $m$ are positive integers between one and $r$ that are selected by the user of our structural result in order to satisfy the four conditions of Theorem~\ref{thm:structural}; the same is true for the values $d_i$, $i=1,\dots, q$. It is precisely this flexibility in the construction of the matrix $\Sreg$ that makes Theorem~\ref{thm:structural} able to accommodate different constructions of the sketching matrix $\Wb$.
\begin{theorem}\label{thm:structural}
Let $\Ab\in\RR{n}{d}$ be the input matrix and let $\Xb\in\RR{d}{(d-k)}$ be any matrix satisfying $\Xb^\ts\Xb = \Ib$, with $1 \leq k < d$. Let the thin SVD of $\Ab$ be $\Ab = \Ub \Sigmab \Vb^\ts$; recall the definition of $\Sreg$ from eqn.~(\ref{eqn:pd1}), and assume that the sketching matrix $\Wb\in \RR{s}{n}$ satisfies the following four conditions for some accuracy parameter $\ve$:
	\begin{align}
	\left\| \Sreg\Ub^\ts \Wb^\ts\Wb \Ub\Sreg - \Sreg^2 \right\|_2
	&\le \ve, \label{eq:struct-1} \\
	\left\| \Sreg\Ub^\ts\Wb^\ts\Wb\Ab_{m,\perp} - \Sreg\Ub^\ts\Ab_{m,\perp} \right\|_F
	&\le \ve \, \|\Ab\Xb\|_F, \label{eq:struct-2} \\
	\nbr{\Ab_{m,\perp}^\ts\Wb^\ts\Wb\Ab_{m,\perp}-\Ab_{m,\perp}^\ts\Ab_{m,\perp}}_F &\le \frac{\ve}{\sqrt{k}}\,\|\Ab\Xb\|_F^2,\ \mbox{and}
	\label{eq:struct-4}\\
	\left|\|\Wb\Ab_{m,\perp}\|_F^2 - \|\Ab_{m,\perp}\|_F^2 \right| &\le \ve\,\|\Ab\Xb\|_F^2
	\label{eq:struct-3} \,.
	\end{align}
Then,
\begin{equation}\label{eqn:conc}
\abs{\|\Wb\Ab\Xb\|_F^2-\|\Ab\Xb\|_F^2}\le\left(d_m^{-2}+2\,d_m^{-1}+2\right) \ve\,\|\Ab\Xb\|_F^2 \,.
\end{equation}
\end{theorem}
\noindent Several comments are necessary to better understand the above structural result. First of all, the four conditions of Theorem~\ref{thm:structural} need to be satisfied for a user-specified matrix $\Sreg$. To be precise, the user of the structural result has the flexibility to choose $q$ (the number of non-zero diagonal entries of $\Sreg$) as well as the values of its entries $d_i$, $i=1,\dots, q$, subject to the constraint that the entries are decreasing and strictly positive. Second, the user of the structural result has the flexibility to choose the parameter $m$ (which ranges between one and $q$) that appears in the last three conditions of Theorem~\ref{thm:structural}. In particular, $m$ is used to define the optimal rank-$m$ approximation $\Ab_m$ to the input matrix $\Ab$ (see Section~\ref{sec:prelim} for notation) and the (perpendicular) matrix $\Ab_{m,\perp}$ which satisfies $\Ab = \Ab_m + \Ab_{m,\perp}$ and $\Ab_m^\ts \Ab_{m,\perp} = \zero$. We emphasize that the conditions of Theorem~\ref{thm:structural} only need to hold for \textit{a single user-specified} choice of $m$ without affecting the generality of the theorem's conclusion. Third, all four conditions of Theorem~\ref{thm:structural} boil down to sketching-based matrix multiplication, as described in Section~\ref{sec:mm}. Fourth, the final error bound depends on the accuracy parameter $\ve$ as well as the (user-specified) value $d_m$. As we will see in two different constructions of the sketching matrix $\Wb$ in Section~\ref{sxn:examples}, $d_m$ is a small constant and thus the term in parentheses in the right-hand side of eqn.~(\ref{eqn:conc}) can be easily replaced by a constant. Fifth, we note that the third constraint is a bit tighter (by a factor of $\nicefrac{1}{\sqrt{k}}$) compared to the other ones. This turns out not to be a problem for known constructions of $\Wb$ since the product $\Ab_{m,\perp}^\ts \Ab_{m, \perp}$ is easy to approximate by sketching. Sixth, more general versions of our structural result are possible: for example, one could remove the assumption that the entries of the diagonal matrix $\Sreg$ are decreasing. A more general result could be derived for the general case $d_i \neq 0$ for all $i=1,\dots, q$. We are not aware of a construction of the sketching matrix $\Wb$ that would necessitate this more general setting and, therefore, we refrain from introducing additional complexity to Theorem~\ref{thm:structural}.

\begin{proof}
Throughout the proof, we will make heavy use of the notation of Section~\ref{sec:prelim}. We also introduce an additional piece of notation, namely the diagonal matrix $\Sreg_{\perp}\in\RR{r}{r}$, which is defined as
\begin{equation}\label{eqn:pd2}
 \Sreg_{\perp}=\diag\{\underbrace{0,\ldots, 0}_{q},\underbrace{1,\ldots,1}_{r-q}\}.
\end{equation}
We note that the inverse of $\Sreg+\Sreg_\perp$ always exists, since it is a diagonal matrix with non-zero entries. For notational convenience, let $\hWb \defeq{\Wb^\ts\Wb - \Ib}$. Using properties of the matrix trace and the SVD of $\Ab$, we can rewrite the quantity that we seek to bound in Theorem~\ref{thm:structural} as
	\begin{align}
	\left| \|\Wb\Ab\Xb\|_F^2 - \|\Ab\Xb\|_F^2 \right|
	& = \left| \Tr{\Xb^\ts\Ab^\ts \left(\Wb^\ts\Wb - \Ib\right) \Ab\Xb} \right|\nonumber\\
	& = \left| \Tr{\Xb^\ts\Ab^\ts \hWb \Ab\Xb} \right|
	= \left| \Tr{\Xb^\ts\Vb\Sigmab\Ub^\ts \hWb \Ub\Sigmab\Vb^\ts\Xb} \right| \nonumber\\
	& =\left| \Tr{\Xb^\ts\Vb(\Sigmab_m+\Sigmab_{m, \perp})\Ub^\ts \hWb \Ub(\Sigmab_m+\Sigmab_{m, \perp})\Vb^\ts\Xb} \right|\nonumber\\
	& \leq  \underbrace{\left| \Tr{\Xb^\ts\Vb\Sigmab_m\Ub^\ts \hWb \Ub\Sigmab_m\Vb^\ts\Xb} \right|}_{\Delta_1}+
	\underbrace{\left| \Tr{\Xb^\ts\Vb\Sigmab_{m,\perp}\Ub^\ts \hWb \Ub\Sigmab_{m,\perp}\Vb^\ts\Xb} \right|}_{\Delta_2}\nonumber\\
	& ~~~~~~~~~~~~~~~~~~~~~~~~~~~~~~~~~~~~~~~ + 2\,\underbrace{\left| \Tr{\Xb^\ts\Vb\Sigmab_m\Ub^\ts \hWb \Ub\Sigmab_{m,\perp}\Vb^\ts\Xb} \right|}_{\Delta_3}\,.\label{eq:decompgen}
	\end{align}
In the above derivations, we also used the fact that $\Sigmab = \Sigmab_m + \Sigmab_{m,\perp}$ (see Section~\ref{sec:prelim}).

\paragraph{Bounding $\Delta_1$.}
We start by bounding the first term in eqn.~\eqref{eq:decompgen}:
	\begin{flalign}
	\Delta_1=&\left| \Tr{\Xb^\ts\Vb\Sigmab_m\Ub^\ts \hWb \Ub\Sigmab_m\Vb^\ts\Xb} \right|\nonumber\\
	=&\left| \Tr{\Xb^\ts\Vb\Sigmab_m\big(\Sreg+\Sreg_\perp\big)^{-1}\,\big(\Sreg+\Sreg_\perp\big)\Ub^\ts \hWb \Ub\big(\Sreg+\Sreg_\perp\big)\,\big(\Sreg+\Sreg_\perp\big)^{-1}\Sigmab_m\Vb^\ts\Xb} \right|\nonumber\\
	=&\left| \Tr{\Xb^\ts\Vb\big(\Sreg+\Sreg_\perp\big)^{-1}\,\Sigmab_m\big(\Sreg+\Sreg_\perp\big)\Ub^\ts \hWb \Ub\big(\Sreg+\Sreg_\perp\big)\,\Sigmab_m\big(\Sreg+\Sreg_\perp\big)^{-1}\Vb^\ts\Xb} \right|\nonumber\\
	=&\left| \Tr{\Xb^\ts\Vb\big(\Sreg+\Sreg_\perp\big)^{-1}\,\Sigmab_m\,\Sreg\,\Ub^\ts \hWb\, \Ub\,\Sreg\,\Sigmab_m\big(\Sreg+\Sreg_\perp\big)^{-1}\Vb^\ts\Xb} \right|\nonumber\\
	=&\left| \Tr{\Xb^\ts\Vb\,\Sigmab_m\big(\Sreg+\Sreg_\perp\big)^{-1}\,\Sreg\,\Ub^\ts \hWb\, \Ub\,\Sreg\,\big(\Sreg+\Sreg_\perp\big)^{-1}\Sigmab_m\,\Vb^\ts\Xb} \right|\nonumber\\
	=&\left| \Tr{\Xb^\ts\Vb\Sigmab_m\big(\Sreg+\Sreg_\perp\big)^{-1} \,\Eb_1\,
		\big(\Sreg+\Sreg_\perp\big)^{-1} \Sigmab_m \Vb^\ts\Xb} \right|\nonumber\\
	=& \left| \Tr{\Eb_1 \left(\big(\Sreg+\Sreg_\perp\big)^{-1} \Sigmab_m \Vb^\ts\Xb\right)
		\left(\big(\Sreg+\Sreg_\perp\big)^{-1} \Sigmab_m \Vb^\ts\Xb\right)^\ts} \right|.\label{eq:delta1_r}
	\end{flalign}
In the above, we set $\Eb_1=\Sreg\Ub^\ts \hWb \Ub\Sreg=\Sreg\Ub^\ts\,\Wb^\ts\Wb\,\Ub\Sreg-\Sreg^2$. Further, we used the fact that $\Sigmab_m\big(\Sreg+\Sreg_\perp\big)=\,\Sigmab_m\,\Sreg$, which follows from $\Sigmab_m\,\Sreg_\perp=\zero$ (recall that $m \leq q$). The last equality follows from the invariance of matrix trace under cyclic permutations.

Next, we apply von Neumann's trace inequality and the condition of eqn.~\eqref{eq:struct-1} on the right hand side of eqn.~\eqref{eq:delta1_r} to get
	\begin{align*}
	\Delta_1
	\le&\,\sum_{i=1}^m \sigma_i(\Eb_1) \cdot
	\sigma_i^2\left( \big(\Sreg+\Sreg_\perp\big)^{-1} \Sigmab_m \Vb^\ts\Xb \right)\nonumber\\
	\le&\,\ve \sum_{i=1}^m \sigma_i^2\left( \big(\Sreg+\Sreg_\perp\big)^{-1} \Sigmab_m \Vb^\ts\Xb \right)\\
	\le&\, \ve \left\| \big(\Sreg+\Sreg_\perp\big)^{-1} \Sigmab_m \Vb^\ts\Xb \right\|_F^2.
	\end{align*}
\noindent Notice that $\big(\Sreg+\Sreg_\perp\big)^{-1}\Sigmab_m  = (\Sreg+\Sreg_\perp)_m^{-1}\Sigmab_m$, where $(\Sreg+\Sreg_\perp)_m^{-1}\in\RR{r}{r}$ is a diagonal matrix whose top $m$ diagonal entries are equal to those of $\big(\Sreg+\Sreg_\perp\big)^{-1}$ and the remaining $r-m$ diagonal entries are set to zero. Then,
\begin{align}
	\left\| \big(\Sreg+\Sreg_\perp\big)^{-1} \Sigmab_m \Vb^\ts\Xb \right\|_F^2
	& = \left\| (\Sreg+\Sreg_\perp)_m^{-1} \Sigmab_m \Vb^\ts\Xb \right\|_F^2\nonumber\\
	& \le \left\|(\Sreg+\Sreg_\perp)_m^{-1} \right\|_2^2 \cdot \left\| \Sigmab_m \Vb^\ts\Xb \right\|_F^2\\
	& \le d_m^{-2} \, \|\Ab_m \Xb\|_F^2\,. \label{eq:Delta1-5}
\end{align}
The last inequality follows from
	$
	\left\|(\Sreg+\Sreg_\perp)_m^{-1} \right\|_2 = d_m^{-1}
	$
and the fact that (see Section~\ref{sec:prelim})
	$$
	\left\| \Sigmab_m \Vb^\ts\Xb \right\|_F^2
	= \left\| \Ub \Sigmab_m \Vb^\ts\Xb \right\|_F^2
	= \|\Ab_m \Xb\|_F^2 \,.
	$$
Therefore, we have shown that
	\begin{align}
	\Delta_1
	\le d_m^{-2}\,\ve\, \|\Ab_m \Xb\|_F^2
	\le d_m^{-2}\,\ve\, \|\Ab \Xb\|_F^2 \,,
	\label{eq:Delta5}
	\end{align}
	where the last inequality follows from
	$\|\Ab \Xb\|_F^2 = \|\Ab_m \Xb\|_F^2 + \|\Ab_{m,\perp} \Xb\|_F^2$
	(by the matrix Pythagorean theorem).

\paragraph{Bounding $\Delta_2$.} We now manipulate the second term of eqn.~\eqref{eq:decompgen}. Let $\Xb_\perp\in\RR{d}{k}$ form a basis for the space that is orthogonal to the space spanned by the columns of $\Xb$ (thus, $\Xb\Xb^\ts + \Xb_\perp \Xb_\perp^\ts = \Ib_d$). Using standard properties of the matrix trace, we obtain
	\begin{flalign}
	\Delta_2
	& = \left| \Tr{\Xb^\ts\Vb\Sigmab_{m,\perp}\Ub^\ts \hWb \Ub\Sigmab_{m,\perp}\Vb^\ts\Xb} \right|
	  = \left| \Tr{\Xb^\ts\Ab_{m,\perp}^\ts \hWb \Ab_{m,\perp}\Xb} \right|\nonumber\\
	& = \left| \Tr{\Ab_{m,\perp}^\ts \hWb \Ab_{m,\perp}\Xb\Xb^\ts} \right|
	  =\left| \Tr{\Ab_{m,\perp}^\ts \hWb \Ab_{m,\perp}(\Ib_d-\Xb_\perp\Xb_\perp^\ts)} \right|\nonumber\\
	& =\left| \Tr{\Ab_{m,\perp}^\ts \hWb \Ab_{m,\perp}-\Ab_{m,\perp}^\ts \hWb \Ab_{m,\perp}\Xb_\perp\Xb_\perp^\ts} \right|\nonumber\\
	& \le \underbrace{\left| \Tr{\Ab_{m,\perp}^\ts \hWb \Ab_{m,\perp}}\right|}_{\Delta_{21}}+\underbrace{\left| \Tr{\Ab_{m,\perp}^\ts \hWb \Ab_{m,\perp}\Xb_\perp\Xb_\perp^\ts}\right|}_{\Delta_{22}}.\label{eq:delta2_r}
	\end{flalign}
We now bound $\Delta_{21}$ and $\Delta_{22}$ separately. Using the structural condition of eqn.~\eqref{eq:struct-3}, we get
	\begin{flalign}
	\Delta_{21}
	& = \left| \Tr{\Ab_{m,\perp}^\ts \hWb \Ab_{m,\perp}}\right|
	= \left| \Tr{\Ab_{m,\perp}^\ts \Wb^\ts\Wb \Ab_{m,\perp}-\Ab_{m,\perp}^\ts \Ab_{m,\perp}}\right|\nonumber\\
	& = \left|\|\Wb\Ab_{m,\perp}\|_F^2-\|\Ab_{m,\perp}\|_F^2\right|
	 \le \ve\,\|\Ab\Xb\|_F^2\,.\label{eq:delta21}
	\end{flalign}
In order to bound $\Delta_{22}$, we will apply \textit{von Neumann}'s trace inequality, the Cauchy-Schwarz inequality, and the structural condition of eqn.~\eqref{eq:struct-4}. For notational convenience, let $\Eb_2=\Ab_{m,\perp}^\ts \hWb \Ab_{m,\perp}=\Ab_{m,\perp}^\ts \Wb^\ts\Wb \Ab_{m,\perp}-\Ab_{m,\perp}^\ts \Ab_{m,\perp}$ and proceed as follows:
\begin{flalign}
	\Delta_{22}
	& = \left| \Tr{\Ab_{m,\perp}^\ts \hWb \Ab_{m,\perp}\Xb_\perp\Xb_\perp^\ts}\right|=\left| \Tr{\Eb_2\Xb_\perp\Xb_\perp^\ts}\right|\nonumber\\
	& \le \sum_{i=1}^k \sigma_i(\Eb_2)\cdot \sigma_i(\Xb_{\perp}\Xb_{\perp}^\ts)
	  \le \left[\sum_{i=1}^k \sigma_i^2(\Eb_2)\right]^{\frac{1}{2}} \cdot \left[\sum_{i=1}^k\sigma_i^2(\Xb_{\perp}\Xb_{\perp}^\ts)\right]^{\frac{1}{2}}\label{eqn:pd3}\\
	& \le \sqrt{k}\nbr{\Eb_2}_F
	  \le \ve\,\|\Ab\Xb\|_F^2 \,.\label{eq:delta22}
\end{flalign}
It is important to note that the matrix $\Xb_\perp$ has rank $k$ and thus has at most $k$ non-zero singular values (all equal to one), which explains the fact that the summation in eqn.~(\ref{eqn:pd3}) stops at $k$. The final two inequalities follow from $\sum_{i=1}^k \sigma_i^2(\Eb_2)\le\nbr{\Eb_2}_F^2$ and the structural condition of eqn.~(\ref{eq:struct-4}). Finally combining eqns.~\eqref{eq:delta2_r}, \eqref{eq:delta21}, and \eqref{eq:delta22} we conclude
\begin{flalign}
	\Delta_2\leq 2\,\ve\,\|\Ab\Xb\|_F^2 \,.\label{eq:delta2}
\end{flalign}

\paragraph{Bounding $\Delta_3$.} Finally, we consider the third term of eqn.~\eqref{eq:decompgen}:
\begin{align}
	\Delta_3
	& =\left| \Tr{\Xb^\ts\Vb\Sigmab_m\Ub^\ts \hWb \,\Ub\Sigmab_{m,\perp}\Vb^\ts\,\Xb} \right|\nonumber\\
	&=\left| \Tr{\Xb^\ts\Vb\Sigmab_m\Ub^\ts \hWb \Ab_{m,\perp}\Xb} \right|\nonumber\\
	&= \left| \Tr{\Xb^\ts\Vb\Sigmab_m\big(\Sreg+\Sreg_\perp\big)^{-1}\big(\Sreg+\Sreg_\perp\big)\Ub^\ts \hWb \Ab_{m,\perp}\Xb} \right|\nonumber\\
	&= \left| \Tr{\Xb^\ts\Vb\big(\Sreg+\Sreg_\perp\big)^{-1}\Sigmab_m\big(\Sreg+\Sreg_\perp\big)\Ub^\ts \hWb \Ab_{m,\perp}\Xb} \right|\nonumber\\
	&= \left| \Tr{\Xb^\ts\Vb\big(\Sreg+\Sreg_\perp\big)^{-1}\Sigmab_m\Sreg\,\Ub^\ts\, \hWb \Ab_{m,\perp}\Xb} \right|\nonumber\\
	&= \left| \Tr{\Xb^\ts\Vb\Sigmab_m\big(\Sreg+\Sreg_\perp\big)^{-1}\Sreg\,\Ub^\ts\, \hWb \Ab_{m,\perp}\Xb} \right|.\label{eq:delta3_r0}
\end{align}
In the above, we repeatedly used the fact that matrix multiplication of diagonal matrices is commutative and $\Sigmab_m\Sreg_\perp = \zero$.
Next, we apply \textit{von Neumann}'s trace inequality and the Cauchy-Schwarz inequality to eqn.~\eqref{eq:delta3_r0} and obtain
	\begin{align}
	\Delta_3
	&\le \sum_{i=1}^m \sigma_i\left( \Xb^\ts\Vb \Sigmab_m\big(\Sreg+\Sreg_\perp\big)^{-1} \right) \cdot
	\sigma_i\left( \Sreg\Ub^\ts\hWb\Ab_{m, \perp}\Xb \right) \nonumber\\
	& \le \bigg[\sum_{i=1}^m \sigma_i^2\left( \Xb^\ts\Vb \Sigmab_m\big(\Sreg+\Sreg_\perp\big)^{-1} \right) \bigg]^{\frac{1}{2}}
	\cdot \bigg[\sum_{i=1}^m \sigma_i^2\left( \Sreg\Ub^\ts\hWb\Ab_{m, \perp}\Xb \right) \bigg]^{\frac{1}{2}} \nonumber\\
	& = \|\Xb^\ts\Vb \Sigmab_m\big(\Sreg+\Sreg_\perp\big)^{-1}\|_F \cdot \|\Sreg\Ub^\ts\hWb\Ab_{m, \perp}\Xb\|_F\nonumber\\
	&\le \|\Xb^\ts\Vb \Sigmab_m\big(\Sreg+\Sreg_\perp\big)^{-1}\|_F \cdot \|\Sreg\Ub^\ts\hWb\Ab_{m, \perp}\|_F \,,\label{eq:delta3_r}
	\end{align}
where the last inequality follows from strong submultiplicativity and the fact that $\|\Xb\|_2=1$. Note that in eqn.~\eqref{eq:Delta1-5} we proved that $\|\Xb^\ts\Vb \Sigmab_m\big(\Sreg+\Sreg_\perp\big)^{-1}\|_F \le d_m^{-1}\|\Ab_m\Xb\|_F$, while the structural condition of eqn.~\eqref{eq:struct-2} gives
\[
\|\Sreg\Ub^\ts\hWb\Ab_{m, \perp}\|_F=\|\Sreg\Ub^\ts\,\Wb^\ts\Wb\,\Ab_{m, \perp}-\Sreg\Ub^\ts\Ab_{m, \perp}\|_F\le\ve\|\Ab\Xb\|_F \,.
\]
Thus, eqn.~\eqref{eq:delta3_r} can be bounded as
\begin{flalign}
\Delta_3\le d_m^{-1} \, \|\Ab_m\Xb\|_F\cdot\ve\|\Ab\Xb\|_F\le\,d_m^{-1}\,\ve\,\|\Ab\Xb\|_F^2 \,,\label{eq:delta3_r2}
\end{flalign}
where the last inequality follows from $\|\Ab\Xb\|_F^2 = \|\Ab_m\Xb\|_F^2 + \|\Ab_{m,\perp}\Xb\|_F^2$
(by the matrix Pythagorean theorem).

\paragraph{Final bound.}
	Combining eqns.~\eqref{eq:decompgen}, \eqref{eq:Delta5}, \eqref{eq:delta2}, and \eqref{eq:delta3_r2} concludes the proof of eqn.~(\ref{eqn:conc}).
\end{proof}


\section{Satisfying the Conditions of Theorem~\ref{thm:structural}}\label{sxn:examples}

In this section we show how to satisfy the conditions of Theorem~\ref{thm:structural} using various constructions for the sketching matrix $\Wb$. Of particular interest is that our structural result of Theorem~\ref{thm:structural} unifies the sketching matrix constructions of~\cite{CohEldMusMusetal15} and~\cite{CohMusMus17}.

\subsection{Review of Randomized Matrix Multiplication}\label{sec:mm}

We present a brief review of randomized matrix multiplication and some relevant theoretical results from prior work that will be useful in this section. Consider a simple algorithm (Algorithm~\ref{algo:1}) to construct a \textit{sampling-and-rescaling} matrix $\Wb \in \mathbb{R}^{s \times n}$.
Using Algorithm~\ref{algo:1} we can approximate the matrix product $\Ab \Bb$ by $\Ab\Wb^\ts \Wb\Bb$, where $\Ab\Wb^\ts$ is the sketch of $\Ab$ and $\Wb\Bb$ is the sketch of $\Bb$.

\begin{algorithm}
    \caption{Construct sampling-and-rescaling matrix}\label{algo:1}
    \begin{algorithmic}
        \STATE \textbf{Input:} Probabilities $p_k$, $k=1,\dots, n$; integer $s \ll n$;
        \STATE \textbf{Initialize:} $\Wb \gets \zero_{s \times n}$;

        \FOR{$i=1$ \textbf{to} $s$}
        \STATE Pick $j_i\in\left\{1,\ldots ,n\right\}$ with $\pr{j_i=k}=p_k$;
        \STATE $\Wb_{ij_i} \gets (s\,p_{j_i})^{-\frac{1}{2}}$;
        \ENDFOR

        \STATE \textbf{Output:} Sampling-and-rescaling matrix $\Wb$;
    \end{algorithmic}
\end{algorithm}

\begin{lemma}\label{lem:exvar}
Given matrices $\Ab\in\RR{m}{n}$ and $\Bb\in\RR{n}{p}$,
let $\Wb\in \RR{s}{n}$ be constructed using Algorithm~\ref{algo:1}. Then,
\begin{equation}
	\Ex{\|\Ab\Wb^\ts\Wb\Bb - \Ab\Bb\|_F^2} \le
	\sum_{i=1}^n \frac{\|\Ab_{*i}\|_2^2 \cdot \|\Bb_{i*}\|_2^2}{sp_i}\label{eq:exf_1}\,.
\end{equation}
Furthermore, if $m = p$,
\begin{equation}
\Ex{\left(\Tr{\Ab\Wb^\ts\Wb\Bb - \Ab\Bb} \right)^2}
\le \sum_{i=1}^n \frac{\left[ (\Bb\Ab)_{ii} \right]^2}{sp_i}\label{eq:extr_1} \,.
\end{equation}
\end{lemma}
\noindent Eqn.~(\ref{eq:exf_1}) was proven in~\cite{DKM06a} (see Lemma 3); the proof of eqn.~(\ref{eq:extr_1}) is a simple exercise using the setup of Lemma 3 in~\cite{DKM06a}.
\begin{lemma}\label{cor:matmul}
Given matrices $\Ab\in\RR{m}{n}$, $\Bb\in\RR{n}{p}$ and for some constant $\beta\in\,(0,1]$, let $\Wb\in \RR{s}{n}$ be constructed using Algorithm~\ref{algo:1} with
$$p_i\ge\beta\frac{\nbr{\Ab_{*i}}_2^2}{\nbr{\Ab}_F^2}\,,\quad \forall i=1,\dots,n$$
such that $\sum_{i=1}^n p_i=1$. Then,
\begin{equation}
	\Ex{\|\Ab\Wb^\ts\Wb\Bb - \Ab\Bb\|_F^2} \le
	\frac{1}{\beta s}\nbr{\Ab}_F^2 \cdot \nbr{\Bb}_F^2\,.
\end{equation}
Furthermore, if $m = p$,
\begin{equation}
	\Ex{\left(\Tr{\Ab\Wb^\ts\Wb\Bb - \Ab\Bb} \right)^2}
	\le \frac{1}{\beta s}\nbr{\Ab}_F^2 \cdot \nbr{\Bb}_F^2.
\end{equation}
\end{lemma}
The proof of the above lemma is immediate from Lemma~\ref{lem:exvar} (with an application of the Cauchy-Schwartz inequality which implies that $\left((\Bb\Ab)_{ii}\right)^2 \leq \|\Ab_{*i}\|_2^2 \cdot \|\Bb_{i*}\|_2^2$). Finally, the next lemma appeared in~\cite{CYD18} as Theorem~3 and is a strengthening of Theorem~4.2 of \cite{Holodnak2015} for the special case when $\nbr{\Ab}_2\le1$. We also note that Lemma~\ref{lem:matmul} is implicit in \cite{CohMusMus17}\,.
\begin{lemma}\label{lem:matmul}
Given $\Ab \in \mathbb{R}^{m \times n}$ with $\nbr{\Ab}_2\leq 1$, let $\Wb\in \RR{s}{n}$ be constructed using Algorithm~\ref{algo:1} with $p_i\ge\beta\nbr{\Ab_{i*}}_2^2/\nbr{\Ab}_F^2$ for all  $i=1,\dots,n$ and $\beta\in\,(0,1]$ such that $\sum_{i=1}^n p_i=1$. Let $\delta$ be a failure probability and $\ve>0$ be an accuracy parameter. If the number of sampled columns $s$ satisfies
\[
s\ge 2\left(1 + \frac{\ve}{3}\right)\frac{\|\Ab\|_F^2}{\beta\,\ve^2}\,\ln\left(\frac{4\,(1+\nbr{\Ab}_F^2)}{\delta}\right),
\]
%
then, with probability at least  $1-\delta$,
\begin{flalign*}
	\nbr{\Ab\Wb^\ts\Wb\Ab-\Ab\Ab^\ts}_2\le\ve.
\end{flalign*}
\end{lemma}

\subsection{Leverage Score-Based Sampling}

Our first approach constructs a \textit{sampling-and-rescaling} matrix $\Wb$ using Algorithm~\ref{algo:1} with sampling probabilities $p_i$, $i=1,\dots,n$:
\begin{equation}\label{eqn:pd11}
p_i \defeq \frac{1}{2} \frac{\|(\Ub_k)_{i*}\|_2^2}{k} + \frac{1}{2} \frac{\|(\Ab_{k,\perp})_{i*}\|_2^2}{\|(\Ab_{k,\perp})\|_F^2} \,.
\end{equation}
Clearly $\sum_{i=1}^n p_i = 1$. Recall that $(\Ub_k)_{i*}$ denotes the $i$-th row of the matrix of the top $k$ left singular vectors of $\Ab$, while $(\Ab_{k,\perp})_{i*}$ denotes the $i$-th row of the matrix $\Ab_{k,\perp} = \Ab - \Ab_k$ (here $\Ab_k$ is the best rank $k$ approximation to $\Ab$). It is well-known that the quantities $\|(\Ub_k)_{i*}\|_2^2$ for $i=1,\dots,n$ correspond to the so-called leverage scores of the best rank $k$ approximation to $\Ab$ (see~\cite{Drineas2016, Mahoney2009} for detailed discussions of the leverage scores and their properties). The sampling probabilities $p_i$ of eqn.~(\ref{eqn:pd11}) are a linear combination of the aforementioned leverage scores and a quantity that depends on the row norms of the residual matrix $\Ab_{k,\perp}$. It is worth noting that, to the best of our knowledge, using only the leverage scores as the sampling probabilities to construct the sampling-and-rescaling matrix $\Wb$ would not suffice to satisfy all conditions of Theorem~\ref{thm:structural}.

We are now ready to apply Theorem~\ref{thm:structural} in order to analyze the performance of the matrix $\Wb$ that is constructed using the above procedure. First, recall that, as users of Theorem~\ref{thm:structural}, we have full control in the construction of the matrix $\Sreg$. Towards that end, let $\Sreg$ be constructed as follows:
\begin{equation}\label{eqn:pd111}
\Sreg=\diag\{\underbrace{1,\ldots,1}_{k},\underbrace{0,\ldots,0}_{r-k}\},
\end{equation}
where both $m$ and $q$ (two parameters associated with the matrix $\Sreg$) are set to be equal to $k$. Trivially, $d_m=1$ by the construction of $\Sreg$. Therefore the constant at the right-hand side of eqn.~(\ref{eqn:conc}) is equal to five.

\paragraph{Satisfying the condition of eqn.~\eqref{eq:struct-1}.}
Using our definition for $\Sreg$, we rewrite the left hand side of the structural eqn.~\eqref{eq:struct-1} as follows:
\begin{flalign}
&\left\| \Sreg\Ub^\ts \Wb^\ts\Wb \Ub\Sreg - \Sreg^2 \right\|_2
=\left\| \begin{pmatrix}
\Ib_k && \mathbf{0}\\
\mathbf{0} && \mathbf{0}
\end{pmatrix}\Ub^\ts \Wb^\ts\Wb \Ub\begin{pmatrix}
\Ib_k && \mathbf{0}\\
\mathbf{0} && \mathbf{0}
\end{pmatrix} - \begin{pmatrix}
\Ib_k && \mathbf{0}\\
\mathbf{0} && \mathbf{0}
\end{pmatrix} \right\|_2\nonumber\\
=&\left\| \begin{pmatrix}
\Ub_k^\ts \Wb^\ts\Wb \Ub_k-\Ib_k && \mathbf{0}\\
\mathbf{0} && \mathbf{0}
\end{pmatrix} \right\|_2
=\left\|\Ub_k^\ts \Wb^\ts\Wb \Ub_k-\Ib_k\right\|_2.\label{eq:struct11}
\end{flalign}

\noindent Notice that the sampling probabilities $p_i$ satisfy $p_i \ge \frac{1}{2} \frac{\|(\Ub_k)_{i*}\|_2^2}{k}$ for $i=1,\dots,n$. Thus, combining eqn.~\eqref{eq:struct11} and Lemma~\ref{lem:matmul}, we get
\begin{equation}\label{eq:bernstein}
\pr{\|\Sreg\Ub^\ts \Wb^\ts \Wb \Ub\Sreg - \Sreg^2\|_2 \ge \ve} \le \frac{\delta}{4} \,.
\end{equation}
For the above bound to hold, we need to set the number of sampled rows of $\Ab$\,,
%

\[
s\ge \left(1 + \frac{\ve}{3}\right)\frac{4\,k\,\ln\left(\nicefrac{16\,(1+k)}{\delta}\right)}{\ve^2}\,.
\]


\paragraph{Satisfying the condition of eqn.~\eqref{eq:struct-2}.}
We start by proving a simple inequality that will be useful in subsequent derivations (recall the definition of $\Xb_{\perp}$ from Section~\ref{sxn:structural}):
\begin{flalign}
\|\Ab_{k,\perp}\|_F
=\|\Ab-\Ab_k\|_F^2\le\|\Ab-\Ab\Xb_{\perp}\Xb_\perp^\ts\|_F
=\|\Ab\Xb\Xb^\ts\|_F
=\|\Ab\Xb\|_F\label{eq:ineq1}.
\end{flalign}
The only inequality in the above derivation is due to the fact that $\Ab_k$ is the \emph{best} rank-$k$ approximation to $\Ab$.
We now use the definition of $\Sreg$ to rewrite the condition as follows:
\begin{flalign}
&\left\|\Sreg\Ub^\ts\Wb^\ts\Wb\Ab_{k,\perp} - \Sreg\Ub^\ts\Ab_{k,\perp} \right\|_F\nonumber\\
=&\left\|\begin{pmatrix}
\Ib_k && \mathbf{0}\\
\mathbf{0} && \mathbf{0}
\end{pmatrix}\begin{pmatrix}
\Ub_k^\ts \\
\Ub_{k,\perp}^\ts
\end{pmatrix}\Wb^\ts\Wb\Ab_{k,\perp} - \begin{pmatrix}
\Ib_k && \mathbf{0}\\
\mathbf{0} && \mathbf{0}
\end{pmatrix}\begin{pmatrix}
\Ub_k^\ts \\
\Ub_{k,\perp}^\ts
\end{pmatrix}\Ab_{k,\perp} \right\|_F\nonumber\\
=&\left\|\begin{pmatrix}
\Ub_k^\ts \\
\zero
\end{pmatrix}\Wb^\ts\Wb\Ab_{k,\perp} - \begin{pmatrix}
\Ub_k^\ts \\
\zero
\end{pmatrix}\Ab_{k,\perp} \right\|_F
= \left\| \Ub_k^\ts\Wb^\ts\Wb\Ab_{k,\perp} - \underbrace{\Ub_k^\ts\Ab_{k,\perp}}_{\zero} \right\|_F\label{eq:delta3_1}.
\end{flalign}
We emphasize that $\Ub_k^\ts\Ab_{k,\perp}=\zero$. Using the fact that $p_i\ge \frac{1}{2} \frac{\|(\Ub_k)_{i*}\|_2^2}{k}$ and applying Lemma~\ref{cor:matmul}, we obtain
\begin{flalign*}
\Ex{\|\Ub_k^\ts\Wb^\ts\Wb\Ab_{k,\perp}\|_F^2}
\le \frac{2}{s}\|\Ub_k\|_F^2\|\Ab_{k, \perp}\|_F^2=\frac{2k}{s}\|\Ab_{k, \perp}\|_F^2\le\frac{2k}{s}\|\Ab\Xb\|_F^2\,.
\end{flalign*}
%
%
where we use the fact $\|\Ub_k\|_F^2=k$ and the last inequality follows from eqn.~\eqref{eq:ineq1}.
Applying Markov's inequality,
we get
\begin{align}
\pr{\|\Ab_{k,\perp}^\ts\Wb^\ts\Wb\Ub_k\|_F \ge \ve\, \|\Ab\Xb\|_F}
& \le \frac{\delta}{4} \,.
\label{eq:markov-3}
\end{align}
The above bound holds if the number of sampled rows
%
$s \ge \nicefrac{8k}{\delta\ve^2}.$

\paragraph{Satisfying the conditions of eqns.~\eqref{eq:struct-4} and~\eqref{eq:struct-3}.}
We note that
$p_i\ge \frac{1}{2} \frac{\|(\Ab_{k,\perp})_{i*}\|_2^2}{\|(\Ab_{k,\perp})\|_F^2}$ for $i=1,\dots,n$. Applying Lemma~\ref{cor:matmul} and using eqn.~(\ref{eq:ineq1}), we obtain
\begin{align*}
\Ex{ \|\Ab_{k,\perp}^\ts\Wb^\ts\Wb\Ab_{k,\perp} - \Ab_{k,\perp}^\ts\Ab_{k,\perp}\|_F^2 }
& \le \frac{2}{s} \, \|\Ab_{k,\perp}\|_F^4\le \frac{2}{s} \, \|\Ab\Xb\|_F^4 \,, \\
\Ex{ \left( \|\Wb\Ab_{k,\perp}\|_F^2 - \|\Ab_{k,\perp}\|_F^2 \right)^2 }
& \le \frac{2}{s} \, \|\Ab_{k,\perp}\|_F^4\le \frac{2}{s} \, \|\Ab\Xb\|_F^4 \,.
\end{align*}
In the above derivations, we used the fact that
\[
\Tr{\Ab_{k,\perp}^\ts\Wb^\ts\Wb\Ab_{k,\perp} - \Ab_{k,\perp}^\ts\Ab_{k,\perp}}
= \|\Wb\Ab_{k,\perp}\|_F^2 - \|\Ab_{k,\perp}\|_F^2 \,.
\]
Next, by Markov's inequality,
we get
\begin{align}
\pr{\left| \|\Wb\Ab_{k,\perp}\|_F^2 - \|\Ab_{k,\perp}\|_F^2 \right|
	\ge \frac{\ve}{\sqrt{k}} \, \|\Ab\Xb\|_F^2}
& \le \frac{\delta}{4}
\label{eq:markov-1} \,, \\
\pr{\|\Ab_{k,\perp}^\ts\Wb^\ts\Wb\Ab_{k,\perp} - \Ab_{k,\perp}^\ts\Ab_{k,\perp}\|_F
	\ge \frac{\ve}{\sqrt{k}} \, \|\Ab\Xb\|_F^2}
& \le \frac{\delta}{4}
\label{eq:markov-2} \,.
\end{align}
For the above bounds to hold, we need to set the number of sampled rows
%
$s \ge \nicefrac{8k}{\delta\ve^2}.$
%
It is worth noting that the bound of eqn.~(\ref{eq:markov-1}) is \textit{stronger} (by a factor of $\nicefrac{1}{\sqrt{k}}$) compared to what is needed in Theorem~\ref{thm:structural}. This improved bound comes for free given the value of $s$ used in the construction of $\Wb$ and does not affect the tightness of the overall bound.

Finally, applying the union bound to
eqns.~\eqref{eq:bernstein},~\eqref{eq:markov-3},~\eqref{eq:markov-1}, and~\eqref{eq:markov-2},
we conclude that if the number of sampled rows $s$ satisfies
\begin{equation*}
s \ge \max\left\{\left(1 + \frac{\ve}{3}\right)\frac{4\,k\,\ln\left(\nicefrac{16\,(1+k)}{\delta}\right)}{\ve^2} \,,
\frac{8k}{\delta\,\ve^2}
\right\} \,,
\end{equation*}
then all four structural conditions of Theorem~\ref{thm:structural} hold with probability at least $1- \delta$. Therefore, the number of sampled rows $s$ is, asymptotically (assuming that $\delta$ is constant),
$s = \Ocal\left(\nicefrac{k \ln k}{\ve^2}\right).$

We conclude this section by noting that a similar proof strategy (using the same construction for the matrix $\Sreg$ of eqn.~(\ref{eqn:pd111})) could also be used to prove that all five constructions of sketching matrices described in Lemma~11 of~\cite{CohEldMusMusetal15} return cost-preserving projections.

\subsection{Ridge Leverage Score Sampling}

Our second approach constructs a \textit{sampling-and-rescaling} matrix $\Wb$ using Algorithm~\ref{algo:1} with sampling probabilities $p_i$ that are proportional to the so-called \textit{ridge leverage scores}~\cite{alaoui2015fast,CohMusMus17} of the rows of the matrix $\Ab$. To properly define the ridge leverage scores of the rows of $\Ab$, we first define the $r \times r$ diagonal matrix $\Sigmab_{\lambda}$ as follows:
\begin{equation}\label{eqn:sigmal}
  \Sigmab_{\lambda} = \diag\left\{\frac{\sigma_1}{\sqrt{\sigma_1^2 + \lambda}},\ldots,\frac{\sigma_r}{\sqrt{\sigma_r^2 + \lambda}}\right\}.
\end{equation}
Recall that $r$ is the rank of the matrix $\Ab$. The $i$-th \emph{row ridge leverage score}, denoted by $\tau_i^\lambda$, of $\Ab$ with respect to the ridge parameter $\lambda>0$ is given by
\begin{flalign}
\tau_i^\lambda\defeq\left(\Ab(\Ab^\ts\Ab+\lambda\Ib_d)^{-1}\Ab^\ts\right)_{ii}=\,\nbr{(\Ub\Sigmab_\lambda)_{i*}}_2^2,
\label{def:reglev}
\end{flalign}
for $i=1,\dots,n$. Recall that $(\Ub\Sigmab_\lambda)_{i*}$ denotes the $i$-th row of the matrix of \textit{all} the left singular vectors of $\Ab$, rescaled by the diagonal entries of $\Sigmab_{\lambda}$. The last equality in eqn.~(\ref{def:reglev}) follows by using the SVD of $\Ab$ and the definition of the matrix $\Sigmab_{\lambda}$. Let $d_{\lambda}$ denote the sum of the ridge leverage scores, \ie,
\begin{equation}\label{eqn:dlambda}
d_\lambda = \sum_{i=1}^n \tau_i^{\lambda} = \sum_{i=1}^n\nbr{(\Ub\Sigmab_\lambda)_{i*}}_2^2 = \nbr{\Ub\Sigmab_\lambda}_F^2 =
\nbr{\Sigmab_\lambda}_F^2.
\end{equation}
The last equality follows by the unitary invariance of the Frobenius norm. We can now define the sampling probabilities $p_i$, $i=1,\dots,n$, as
\begin{equation}\label{eqn:prl}
p_i \defeq \frac{\tau_i^\lambda}{\sum_{i=1}^n \tau_i^\lambda} = \frac{\tau_i^\lambda}{d_\lambda} =
\frac{\nbr{(\Ub\Sigmab_\lambda)_{i*}}_2^2}{\nbr{\Sigmab_\lambda}_F^2} \,.
\end{equation}
Clearly $\sum_{i=1}^n p_i = 1$. In the remainder of this section, we will analyze the special case where
\begin{equation}\label{eqn:lambdaval}
\lambda=\frac{\|\Ab_{k,\perp}\|_F^2}{k} \,.
\end{equation}
This is the case analyzed in~\cite{CohMusMus17} and the simplest known setting for $\lambda$ that returns provably accurate approximations for cost-preserving projections via ridge-leverage score sampling.

We now proceed to apply Theorem~\ref{thm:structural} in order to analyze the performance of the matrix $\Wb$ that is constructed using the the ridge leverage scores as sampling probabilities. Recall that, as users of Theorem~\ref{thm:structural}, we have full control of the construction of the matrix $\Sreg$. Towards that end, let
%
\begin{equation}\label{eqn:pd112}
\Sreg=\Sigmab_\lambda.
\end{equation}
For the parameters associated with $\Sreg$ in eqn.~\eqref{eqn:pd1},
we will set $q$ to $r$, the rank of the matrix $\Ab$; and set $m$
to be the index of smallest non-zero singular value of $\Ab$ such that
\begin{equation}\label{eqn:definem}
  \sigma_{m}^2\ge \lambda \ge\sigma_{m+1}^2.
\end{equation}
Several observations follow from the above definitions. First of all, the diagonal entries of $\Sreg$ (denoted as $d_i$ in eqn.~(\ref{eqn:pd1})) are set to $d_i=\nicefrac{\sigma_i}{\sqrt{\sigma_i^2+\lambda}}$. We can upper-bound $d_m^{-1}$ as follows:
\begin{flalign*}
d_m^{-1}=\sqrt{1+\frac{\lambda}{\sigma_{m}^2}}\le\sqrt{2}\,,
\end{flalign*}
where the last inequality follows from our choice for $m$ in eqn.~(\ref{eqn:definem}). This implies that the constant in the right-hand side of eqn.~(\ref{eqn:conc}) is at most $4+2\sqrt{2}$. Second, using our choices for $m$ (eqn.~(\ref{eqn:definem})) and $\lambda$ (eqn.~(\ref{eqn:lambdaval})), we get
\begin{flalign*}
\nbr{\Ab_{k, \perp}}_F^2 = k\lambda \ge k\sigma_{m+1}^2 \ge \sigma_{m+1}^2+\sigma_{m+2}^2+\dots\sigma_{m+k}^2 \,.
\end{flalign*}
Adding $\nbr{\Ab_{k, \perp}}_F^2$ on both sides of the above inequality yields
\begin{flalign}
2\nbr{\Ab_{k, \perp}}_F^2
& \ge (\sigma_{m+1}^2+\sigma_{m+2}^2+\dots\sigma_{m+k}^2)+(\sigma_{k+1}^2+\sigma_{k+2}^2+\dots\sigma_{r}^2)\nonumber\\
& \ge \sigma_{m+1}^2+\sigma_{m+2}^2+\dots\sigma_{r}^2
= \nbr{\Ab_{m,\perp}}_F^2,\label{defm}
\end{flalign}
where the last inequality holds because $m+k\ge k+1$. Combining eqns.~\eqref{eq:ineq1} and~\eqref{defm} results in the following inequality, which will be quite useful in this section:
\begin{flalign}\label{eqn:Amperp}
\nbr{\Ab_{m,\perp}}_F^2\le 2\nbr{\Ab_{k, \perp}}_F^2\le 2\nbr{\Ab\Xb}_F^2.
\end{flalign}
Third, we can upper-bound the sum of the ridge leverage scores (denoted as $d_\lambda$ in eqn.~(\ref{eqn:dlambda})) as follows:
\begin{flalign}
d_\lambda
& = \nbr{\Sigmab_\lambda}_F^2=\sum_{i=1}^r\frac{\sigma_i^2}{\sigma_i^2+\lambda}
=\sum_{i=1}^k\frac{\sigma_i^2}{\sigma_i^2+\lambda}+\sum_{i=k+1}^r\frac{\sigma_i^2}{\sigma_i^2+\lambda}\nonumber\\
& \le k+\sum_{i=k+1}^r\frac{\sigma_i^2}{\lambda}
 =k+\frac{\nbr{\Ab_{k,\perp}}_F^2}{\lambda} = k+k = 2k.\label{eq:df}
\end{flalign}
In the above, we upper-bounded the top $k$ diagonal entries of the matrix $\Sigmab_{\lambda}$ (squared) by one, while we upper-bounded the bottom $r-k$ ones by $\sigma_i^2/\lambda$. To conclude, we used our specific choice for $\lambda$ from eqn.~(\ref{eqn:lambdaval}). Again, this upper-bound for $d_\lambda$ will be useful later in this section.

\paragraph{Satisfying the condition of eqn.~\eqref{eq:struct-1}.} Applying Lemma~\ref{lem:matmul} and setting $s$, the number of sampled rows, to be at least
\[
s\ge \left(1 + \frac{\ve}{3}\right)\frac{4\,k\,\ln\left(\nicefrac{16\,(1+2k)}{\delta}\right)}{\ve^2}\,,
\]
%
we obtain
\begin{equation}\label{eq:bernstein2}
\pr{\|\Sigmab_{\lambda}\Ub^\ts \Wb^\ts \Wb \Ub\Sigmab_{\lambda} - \Sigmab_{\lambda}^2\|_2 \ge \ve} \le \frac{\delta}{4} \,.
\end{equation}
In the above we used $d_{\lambda}\leq 2k$ from eqn.~(\ref{eq:df}).

\paragraph{Satisfying the condition of eqn.~\eqref{eq:struct-2}.} Applying Lemma~\ref{cor:matmul}, we get
\begin{flalign}
\Ex{\left\|\Sigmab_{\lambda}\Ub^\ts\Wb^\ts\Wb\Ab_{m,\perp} - \Sigmab_{\lambda}\Ub^\ts\Ab_{m,\perp} \right\|_F^2}
& \le \frac{1}{s}\nbr{\Sigmab_{\lambda}\Ub^\ts}_F^2 \cdot \nbr{\Ab_{m,\perp}}_F^2\nonumber\\
& = \frac{d_\lambda}{s}\nbr{\Ab_{m,\perp}}_F^2\le\frac{4k}{s}\nbr{\Ab\Xb}_F^2\label{st2}.
\end{flalign}
The last inequality follows since $d_{\lambda}=\|\Sigmab_{\lambda}\Ub^\ts\|_F^2$ is upper-bounded by $2k$ from eqn.~(\ref{eq:df}); we also used eqn.~(\ref{eqn:Amperp}). Next, applying Markov's inequality,
if the number of sampled rows
%
$s\ge\nicefrac{16\,k}{\delta\ve^2}$,
then
%
\begin{flalign}
\pr{\left\| \Sigmab_{\lambda}\Ub^\ts\Wb^\ts\Wb\Ab_{m,\perp} - \Sigmab_{\lambda}\Ub^\ts\Ab_{m,\perp} \right\|_F> \ve \, \|\Ab\Xb\|_F}
\le\frac{\delta}{4}\label{eq:markov-6} \,.
\end{flalign}

\paragraph{Satisfying the conditions of eqns.~\eqref{eq:struct-4} and~\eqref{eq:struct-3}.} We start with eqn.~\eqref{eq:struct-3}. Using standard properties of the trace, we get
\begin{flalign}
\|\Wb\Ab_{m,\perp}\|_F^2
& =\|\Wb\Ub\Sigmab_{m,\perp}\|_F^2
  =\Tr{\Sigmab_{m,\perp}\Ub^\ts\Wb^\ts\Wb\Ub\Sigmab_{m,\perp}}\nonumber\\
& =\Tr{\Sigmab_{m,\perp}\Sigmab_{\lambda}^{-1}\Sigmab_{\lambda}\Ub^\ts\Wb^\ts\Wb\Ub\Sigmab_{m,\perp}}\nonumber\\
& =\Tr{\Sigmab_{\lambda}\Ub^\ts\Wb^\ts\Wb\Ub\Sigmab_{m,\perp}^2\Sigmab_{\lambda}^{-1}}.\label{st3_1}
\end{flalign}
Similarly,
\begin{flalign}
\|\Ab_{m,\perp}\|_F^2=\Tr{\Sigmab_{\lambda}\Ub^\ts\Ub\Sigmab_{m,\perp}^2\Sigmab_{\lambda}^{-1}}.\label{st3_2}
\end{flalign}
Combining Lemma~\ref{cor:matmul} with eqns.~\eqref{st3_1} and~\eqref{st3_2}, we have
\begin{flalign}
\Ex{\left( \|\Wb\Ab_{m,\perp}\|_F^2 - \|\Ab_{m,\perp}\|_F^2 \right)^2}
& = \Ex{\left( \tr(\Sigmab_{\lambda}\Ub^\ts\Wb^\ts\Wb\Ub\Sigmab_{m,\perp}^2\Sigmab_{\lambda}^{-1}-\Sigmab_{\lambda}\Ub^\ts\Ub\Sigmab_{m,\perp}^2\Sigmab_{\lambda}^{-1}) \right)^2} \nonumber\\
& \le \frac{1}{s}\, \|\Sigmab_{\lambda}\Ub^\ts\|_F^2 \cdot \|\Ub\Sigmab_{m,\perp}^2\Sigmab_{\lambda}^{-1}\|_F^2
= \frac{d_\lambda}{s}\, \|\Sigmab_{m,\perp}^2\Sigmab_{\lambda}^{-1}\|_F^2\nonumber\\
& \le \frac{2k}{s}\, \|\Sigmab_{m,\perp}\|_F^2 \cdot \|\Sigmab_{m,\perp}\Sigmab_{\lambda}^{-1}\|_2^2\label{st3_4}\,,
\end{flalign}
where the last inequality follows by strong submultiplicativity and eqn.~(\ref{eq:df}). Note that $\Sigmab_{m,\perp}\Sigmab_{\lambda}^{-1}$ is a diagonal matrix whose $i$-th diagonal entry is equal to
\[(\Sigmab_{m,\perp}\Sigmab_{\lambda}^{-1})_{ii}= \begin{cases}
0, & i\le m\,; \\
\sqrt{\sigma_i^2+\lambda}, & i\geq m+1\,.
\end{cases}
\]
It now follows that
$$\|\Sigmab_{m,\perp}\Sigmab_{\lambda}^{-1}\|_2=\sqrt{\sigma_{m+1}^2+\lambda}\le\sqrt{2\lambda}=\sqrt{\frac{2}{k}}\|\Ab_{k, \perp}\|_F \,,$$
where the inequality follows from eqn.~(\ref{eqn:definem}) and the last equality follows from eqn.~(\ref{eqn:lambdaval}).
In addition, eqn.~\eqref{eqn:Amperp} yields
$$\|\Sigmab_{m,\perp}\|_F^2=\|\Ab_{m,\perp}\|_F^2\le2\|\Ab\Xb\|_F^2 \,,$$
and thus we get,
\begin{flalign}
\Ex{\left( \|\Wb\Ab_{m,\perp}\|_F^2 - \|\Ab_{m,\perp}\|_F^2 \right)^2}
\le \frac{2k}{s}\cdot 2\|\Ab\Xb\|_F^2 \cdot \frac{2\|\Ab_{k, \perp}\|_F^2}{k}
\le\frac{8}{s}\,\|\Ab\Xb\|_F^4 \,,\label{eq:st3_5}
\end{flalign}
where the last inequality follows from eqn.~(\ref{eq:ineq1}). Using similar algebraic manipulations and Lemma~\ref{cor:matmul}, we obtain
\begin{flalign}
\EE\left(\nbr{\Ab_{m,\perp}^\ts\Wb^\ts\Wb\Ab_{m,\perp}-\Ab_{m,\perp}^\ts\Ab_{m,\perp}}_F^2\right)\le\,\frac{8}{s}\,\|\Ab\Xb\|_F^4 \,.\label{eq:st3_6}
\end{flalign}
Next, applying Markov's inequality and setting the number of sampled rows
%
$s\ge\nicefrac{32k}{\delta \ve^2}$,
we get
%
\begin{align}
\pr{\left| \|\Wb\Ab_{m,\perp}\|_F^2 - \|\Ab_{m,\perp}\|_F^2 \right|
	\ge \frac{\ve}{\sqrt{k}} \, \|\Ab\Xb\|_F^2}
& \le \frac{\delta}{4}
\label{eq:markov-4} \,, \\
\pr{\|\Ab_{m,\perp}^\ts\Wb^\ts\Wb\Ab_{m,\perp} - \Ab_{m,\perp}^\ts\Ab_{m,\perp}\|_F
	\ge \frac{\ve}{\sqrt{k}} \, \|\Ab\Xb\|_F^2}
& \le \frac{\delta}{4}
\label{eq:markov-5} \,.
\end{align}
It is worth noting that the bound of eqn.~(\ref{eq:markov-4}) is \textit{stronger} (by a factor of $\nicefrac{1}{\sqrt{k}}$) compared to what is needed in Theorem~\ref{thm:structural}. This improved bound comes for free given the value of $s$ used in the construction of $\Wb$ and does not affect the tightness of the overall bound.

Finally, applying the union bound to eqns.~\eqref{eq:bernstein2},~\eqref{eq:markov-6},~\eqref{eq:markov-4}, and~\eqref{eq:markov-5}, we observe that if the number of sampled rows
%
\begin{equation*}
s \ge \max\left\{\left(1 + \frac{\ve}{3}\right)\frac{4\,k\,\ln\left(\nicefrac{16\,(1+2k)}{\delta}\right)}{\ve^2} \,,
\frac{32k}{\delta \,\ve^2}\right\},
\end{equation*}
%
%
then all four structural conditions of Theorem~\ref{thm:structural} hold with probability at least $1- \delta$. Therefore, the number of sampled rows $s$ is, asymptotically (assuming that $\delta$ is constant),
$s = \Ocal\left(\nicefrac{k \ln k}{\ve^2}\right).$


\section{Conclusion}

Building upon the definition of cost-preserving projections, we have presented a simple structural result connecting the construction of projection-cost preserving sketches to sketching-based matrix multiplication. Our work unifies and generalizes prior known constructions for projection-cost preserving sketches based on (variants of) the standard leverage scores, ridge leverage scores, as well as other constructions.

An interesting open problem would be to understand whether similar structural results for cost-preserving projections can be derived for other Schatten $p$-norms, \eg, for the Schatten infinity norm, which corresponds to the well-known matrix 2-norm. Preliminary work in this direction includes Lemma 26 in~\cite{CohEldMusMusetal15}; to the best of our knowledge other Schatten $p$-norms have not been studied in prior work. Additionally, it would be interesting to study alternative sets of structural conditions that guarantee cost-preserving projections, with the end goal of fully characterizing the problem by presenting both necessary and sufficient conditions for cost-preserving projections for various Schatten $p$-norms.


\subsection*{Acknowledgements}

AC and PD were partially supported by NSF IIS-1661760 and IIS-1661756.
JY was supported by NSF IIS-1149789 and IIS-1618690.


\setlength{\bibsep}{6pt}
\bibliographystyle{plain}
\bibliography{paper}

\end{document}